\documentclass[lettersize,journal]{IEEEtran}
\usepackage{amsmath,amsfonts}
\usepackage{algorithmic}
\usepackage{algorithm}
\usepackage{array}
\usepackage{textcomp}
\usepackage{stfloats}
\usepackage{url}
\usepackage{verbatim}
\usepackage{graphicx}
\usepackage{tabularx}
\usepackage{colortbl}
\usepackage[table]{xcolor}
\usepackage{multirow}
\usepackage{multicol}
\usepackage{booktabs}
\usepackage{pifont}
\usepackage{subcaption}
\newcommand{\cmark}{\ding{51}} %
\newcommand{\xmark}{\ding{55}} %
\usepackage{cite}
\DeclareRobustCommand*{\IEEEauthorrefmark}[1]{%
    \raisebox{0pt}[0pt][0pt]{\textsuperscript{\footnotesize\ensuremath{#1}}}}

\begin{document}

\title{ COMUNI: Decomposing Common and Unique Video Signals for Diffusion-based Video Generation }

\author{
\IEEEauthorblockN{
Mingzhen Sun\IEEEauthorrefmark{1,2},
Weining Wang\IEEEauthorrefmark{1},
Xinxin Zhu\IEEEauthorrefmark{1}, and
Jing Liu\IEEEauthorrefmark{1,2}} \\
\IEEEauthorblockA{\IEEEauthorrefmark{1} Institute of Automation, Chinese Academy of Sciences, Beijing, China} \\
\IEEEauthorblockA{\IEEEauthorrefmark{2}School of Artificial Intelligence, University of Chinese Academy of Sciences, Beijing, China} \\
\IEEEauthorblockA{sunmingzhen2020@ia.ac.cn, \{weining.wang, xinxin.zhu\}@nlpr.ia.ac.cn} \\
\IEEEauthorblockA{Corresponding Author: Jing Liu \quad Email: jliu@nlpr.ia.ac.cn}
}



\maketitle

\begin{abstract}
Since videos record objects moving coherently, adjacent video frames have commonness (similar object appearances) and uniqueness (slightly changed postures). 
To prevent redundant modeling of common video signals, we propose a novel diffusion-based framework, named COMUNI, which decomposes the COMmon and UNIque video signals to enable efficient video generation.
Our approach separates the decomposition of video signals from the task of video generation, thus reducing the computation complexity of generative models.
In particular, we introduce CU-VAE to decompose video signals and encode them into latent features. 
To train CU-VAE in a self-supervised manner, we employ a cascading merge module to reconstitute video signals and a time-agnostic video decoder to reconstruct video frames.
Then we propose CU-LDM to model latent features for video generation, which adopts two specific diffusion streams to simultaneously model the common and unique latent features.
We further utilize additional joint modules for cross modeling of the common and unique latent features, and a novel position embedding method to ensure the content consistency and motion coherence of generated videos.
The position embedding method incorporates spatial and temporal absolute position information into the joint modules. 
Extensive experiments demonstrate the necessity of decomposing common and unique video signals for video generation and the effectiveness and efficiency of our proposed method\footnote{Our codes will be released as in \url{https://anonymous.4open.science/r/COMUNI}.}.
\end{abstract}


\begin{figure*}
\includegraphics[width=\textwidth]{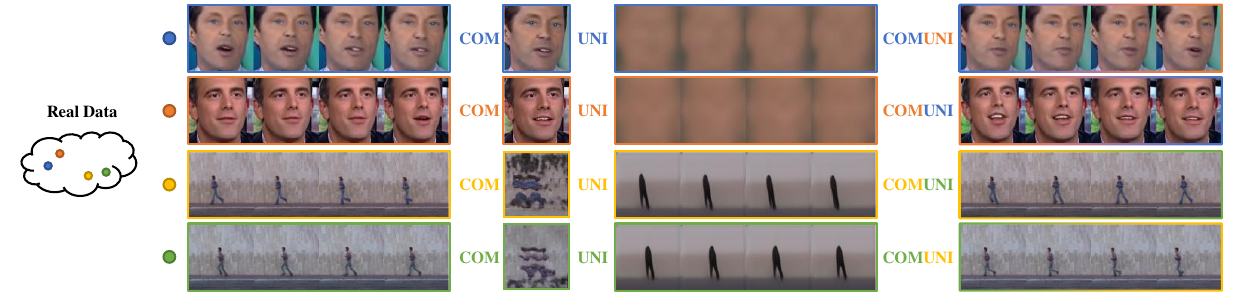}
\caption{
Visualization of the decomposed common and unique video signals.
The first column displays the real videos.
The second column depicts the decomposed common video signal for each video, e.g. human characteristics and video backgrounds.
The third column depicts decomposed unique video signals, e.g. changes of human expressions or body positions, which are in one-to-one correspondence with real video frames.
The last column shows videos decoded from swapped common and unique video signals of two neighbor videos, demonstrating that our method could successfully decompose common and unique video signals and flexibly recompose them when decoding videos.
}
 \label{fig:intro}
\end{figure*}

\section{Introduction}
\IEEEPARstart{D}{iffusion} Probabilistic Models (DPMs) \cite{ddpm, beat, imagen} have shown superior performance in image generation tasks compared to Generative Adversarial Networks (GANs) \cite{stylegan2, projectedGAN} and Auto-Regressive Models (ARMs) \cite{VQGAN, dalle}.
Latent Diffusion Models (LDMs) \cite{ldm}, which utilize a VAE model to encode images as latent features and a diffusion-based generative model to synthesize images, have demonstrated impressive quality in open-domain high-resolution image generation.
However, video generation is more challenging due to the additional temporal dimension, which increases the search space and computational complexity.
Considering that adjacent video frames share both common and unique characteristics, 
we explore whether common and unique video signals can be decomposed and encoded separately and modeled jointly.
Then the redundant modeling of common video signals can be avoided and efficient video generation will be possible.
Moreover, we pursue separating the decomposition of video signals from the task of video generation, which can let generative models focus on modeling video content rather than video details, thereby reducing the computation burden.

To this end, we present COMUNI, a novel two-stage framework that utilizes a VAE model to decompose video signals and a diffusion-based generative model to generate videos.
In the first stage, we explore how to decompose common and unique video signals from a given video clip and encode them into latent features.
To accomplish this, we devise two specific video encoders: a commonness encoder and a uniqueness encoder.
We impose one explicit and one implicit constraint based on the properties of common and unique video signals.
Specifically, we explicitly constrain the commonness encoder to produce a single common feature for the entire video and the uniqueness encoder to create a specific unique feature for each video frame, since common signals are shared among video frames while unique signals are specific to each video frame. 
To prevent unique features from containing common information, we implicitly constrain the uniqueness encoder by setting the spatial shape of unique features to be moderately small.
In this way, the common feature has to capture as much common information as possible, and the unique features have to extract as much unique information as possible to cover more video information (i.e. recover more video details when reconstructing input videos).
For self-supervised training with the target of video reconstruction, we use a cascading merge module to recompose common and unique video signals by fusing corresponding latent features in a cascading manner.
Based on the fused frame-wise video features, a video decoder is used to recover spatial resolution and reconstruct video details.
The two encoders, the cascading merge module and the video decoder make up the Commonness and Uniqueness decomposition VAE model (CU-VAE).
As shown in Fig. \ref{fig:intro}, CU-VAE effectively decomposes common and unique video signals and performs recomposition flexibly.

In the second stage, we propose the Commonness and Uniqueness Latent Diffusion Model (CU-LDM) to generate videos by modeling latent features.
CU-LDM employs two diffusion streams to model common and unique latent features simultaneously.
To obtain consistent generation of common and unique features, we employ multiple joint modules that cross model them.
To reduce computation complexity, the joint modules spatially divide intermediate common and unique features into several blocks and calculate temporal-spatial attention on matched blocks.
Moreover, we propose a novel position embedding method that incorporates absolute position information and enables attention calculation to distinguish between common and unique features.
Specifically, learnable temporal position embeddings are defined respectively for common and unique features, while learnable spatial position embeddings are shared for both.
In particular, since each video is encoded into one common feature and multiple consecutive unique features, a specific temporal embedding is defined for common features to distinguish them from unique features, and a sequence of temporal embeddings is defined to unique features with temporal alignment to help capture temporal relationships.
The spatial position embeddings incorporate information of absolute spatial positions to maintain spatial relationships.
Each temporal position embedding is merged with all spatial position embeddings and is employed without corrupting input features.

Our contributions are as follows:

$\bullet$ We propose a diffusion-based framework COMUNI for efficient video generation, which decomposes videos signals to eliminate redundancy and separates the decomposition of video signals from the task of video generation to reduce computation burden.

$\bullet$ CU-VAE is used to extract common and unique video signals from a video clip and encode them to latent features, which can be recomposed and decoded flexibly.

$\bullet$ CU-LDM is used to model latent features for generation with two diffusion streams and joint modules, and a novel position embedding method is proposed to incorporate absolute position information and differentiate between common and unique latent features.

$\bullet$ Extensive experiments demonstrate the effectiveness and efficiency of our proposed method on multiple benchmarks, such as FaceForensics \cite{ffs} and UCF-101 \cite{ucf}.

\section{Related Work}
\paragraph{Video Generative Models}
Generating videos is a challenging task due to the spatio-temporal complexity and consistency of videos. 
Current mainstream video generation works fall into into three categories: ARMs , GANs, and DPMs.
ARMs \cite{cogvideo,maskvit,videogpt} typically encode videos into discrete tokens and model them with an auto-regressive transformer.
VideoGPT \cite{videogpt} explores the influence of different spatial-temporal downsample factors for auto-regressive video generation.
SVG \cite{svg} proposed a novel Transformer-based generator for auto-regressive sounding video generation.
MOSO \cite{moso} proposed to decompose video motion, scene and object information for video prediction.
Generative Adversarial Networks (GANs) \cite{digan,mocogan,tmmgan2,tmmgan1} usually involves a video generator and a video discriminator, which are trained in an adversarial manner.
DIGAN \cite{digan} employs implicit neural representations in video generation.
MoCoGAN \cite{mocogan} proposes to decouple video signals into content and motion.
G$^3$AN \cite{g3an} introduces a three-stream generator to model the generation of video appearance and motion in a disentangled manner.
However, the generation process of MoCoGAN and G$^3$AN is invertible, and its decoupled video signals are difficult to visualize, which is opposite to our method.

\begin{figure*}
    \centering
    \includegraphics[width=1\linewidth]{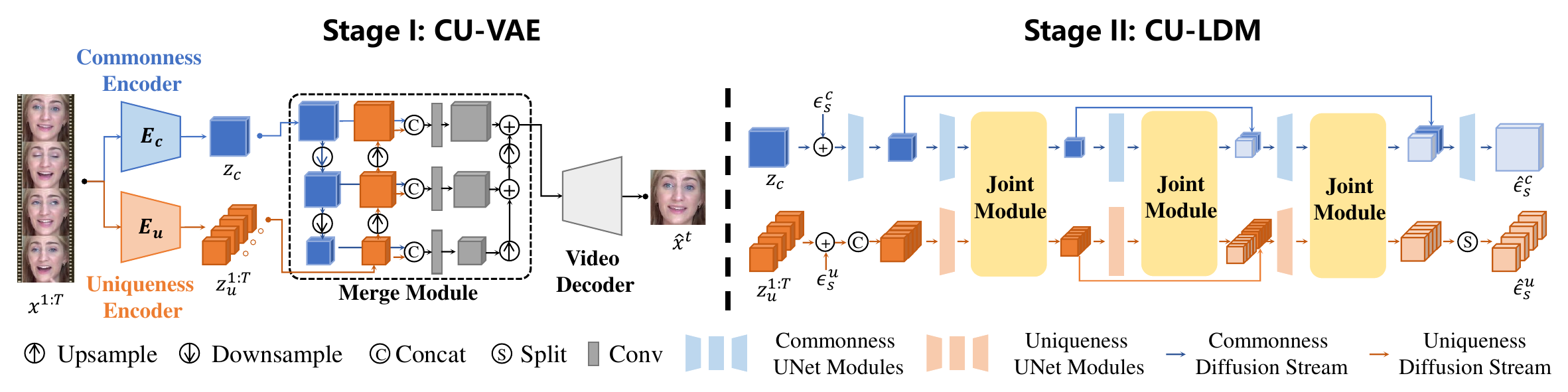}
    \caption{
    The overall architecture of the proposed two-stage framework COMUNI. 
    During the first stage, CU-VAE decomposes common and unique video signals by extracting corresponding information and encoding it into latent features using two specific encoders: the commonness and uniqueness encoders. 
    The merge module then recomposes these features in a cascading manner.
    Based on each fused feature, we adopt a time-agnostic video decoder to reconstruct corresponding video frame.
    In the second stage, CU-LDM employs two diffusion streams to model the common and unique latent features simultaneously.
    Multiple joint modules are interpolated to facilitate cross-modeling of different latent features.
    }
    \label{fig:framework}
\end{figure*}

\paragraph{Diffusion Probabilistic Models}
DPMs have achieved start-of-the-art performance in image generation tasks \cite{nonequilibrium,ddpm,ddim,beat,ldm,imagen}.
Diffusion \cite{nonequilibrium} and DDPM \cite{ddpm} are the pioneering works that utilize diffusion probabilistic models for image generation.
ADM \cite{beat} improved upon the diffusion probabilistic model and achieved better generation performance than GAN-based models.
To address the challenging of high computation consumption when modeling high-resolution images, LDM \cite{ldm} utilizes a KL-VAE to compress images into low-resolution latent features, which can be easily decoded to images.
The method then proceeds to learn the latent features for video generation.
Several works \cite{imagen,unclip,ldm} have achieved unprecedented performance in text-to-image generation using large pretraining models.

However, video generation has seen few advancements since the additional temporal dimension requires models to learn temporal consistency, which significantly increases computation complexity.
VDM \cite{vdm} extended diffusion models to video generation for the first time by extending the 2D backbone to 3D.
FDM \cite{flexible} explored generating a subset of video frames based on another subset of video frames.
MCVD \cite{mcvd} proposed a masked training method to model video prediction, generation, and interpolation tasks simultaneously.
Despite considerable effort, the performance of these methods is still below expectations. 
Large pretraining models \cite{imagenvideo,makeavideo} have achieved impressive quality by stacking multiple specific models, which requires enormous computational resources.

\paragraph{Motion-Content Decomposed Video Generation using DPMs}
To reduce computation complexity in modeling video generation using DPMs, researchers have recently explored decomposing video content and motion information to reduce redundant modeling of common information.
VIDM \cite{vidm} synthesizes an initial video frame as video content using a pretrained image DPM \cite{ldm}.
Then, it generates video motion by auto-regressively predicting subsequent video frames with the assistance of optical-flow \cite{flow}.
Notably, VIDM relies on a well pretrained optical flow estimation network \cite{flow} and requires additional improvements to overcome convergence issues during training \cite{vidm}.
In contrast, our COMUNI does not require auxiliary models like optical-flow networks, and our training is straightforward and easy to converge.

VideoFusion \cite{videofusion} decomposes video content and motion by dividing video noise into shared and residual components.
However, it implicitly performs this decomposition by adjusting the proportion of shared noise during video generation.
In contrast, COMUNI takes an explicit approach to separate video decomposition.
This explicit separation allows us to visualize decomposed common and unique video signals, thereby enhancing model interpretability.

LFDM \cite{cI2V} introduced a novel latent flow diffusion model for the conditional image-to-video generation task.
LaMD \cite{lamd} proposed a latent motion diffusion framework that splits the video generation task into two subtasks: the latent motion generation and video reconstruction subtasks.
Both LFDM and LaMD target on image-to-video and conditional image-to-video generation tasks, which alleviate video generative models from the challenge of synthesizing realistic object appearances and video scenes. However, they necessitate an additional image generative model to produce an initial video frame, and the quality of the final generated video is closely tied to the capabilities of both the image and video generative models. Different from LFDM \cite{cI2V} and LaMD \cite{lamd}, COMUNI is designed for unconditional and text-conditioned video generation tasks. 
This difference allows COMUNI to synthesize both realistic video content and motion simultaneously, eliminating the need for an additional image generative model.

\paragraph{Position Embedding} 
To incorporate positional information, \cite{attention} obtains deterministic position embeddings based on trigonometric functions.
Other works \cite{swin,beit} employ a set of learnable embeddings to capture positional representations. 
To enhance the model understanding of global semantic, these approaches directly add position embeddings to input features for vision understanding tasks.
However, for vision generation tasks, capturing detailed content is much more important than understanding global semantics.
As a result, directly adding position embeddings into input features may lead to confusion for DPMs, which synthesize details of vision content by eliminating noise given a noisy input, about whether or not to treat the added position embeddings as noise.
\cite{relativepe} proposed to add position embedding to key and value features during the calculation of attention layer.
Nonetheless, this approach targets language processing tasks and focuses on incorporating relative position information. 
In this paper, we present a novel position method that incorporates absolute position information when generating videos, which is better suited to diffusion-based generative models.

\section{Method}
We denote a $T$-frame video clip as $x^{1:T}$, where $x^t \in R^{H \times W \times C}$ denotes the $t$-th video frame, with $H$ and $W$ representing the height and width of the video frames, respectively, while $C$ denotes the number of channels.
The overall structure of our two-stage framework, i.e. COMUNI, is illustrated in Fig. \ref{fig:framework}.
In the first stage, CU-VAE decomposes an input video clip into common and unique video signals and encodes them into latent features.
In the second stage, CU-LDM models the common and unique latent features to generate videos.
In this section, we will present our method in details.

\subsection{Stage I: CU-VAE}
\label{sec:cu-vae}
CU-VAE consists of a commonness video encoder, a uniqueness video encoder, a cascading merge module, and a video decoder.
The commonness and uniqueness videos encoders decompose the video signals of an input video clip by extracting the common and unique video signals respectively.
Then, a cascading merge module is employed to recompose video signals, and a video decoder is utilized to reconstruct the input video clip.
Thus, CU-VAE can be trained in a self-supervised manner with the target of video reconstruction.

Although the specific scopes of the common or unique signals are not defined, when different content features are recomposed with the same unique feature, the decoded video frames exhibit distinct object appearances while maintaining the same motion, as depicted in the last column of Fig. 1. 
This result indicates that our video encoders can autonomously learn to extract common or unique information.

\paragraph{Video Decomposition}
For video decomposition, the commonness video encoder $E_c$ and the uniqueness video encoder $E_u$ process the input video clip $x^{1:T}$ to extract its common and unique video signals respectively based on the distinct properties of these signals.
In particular, since common video signals are shared across adjacent video frames, $E_c$ first obtains the frame-wise video features and then compresses their temporal dimension to obtain the spatial commonness of these features.
In this way, we obtain a common latent feature $z_c \in R^{\frac{H}{f_c} \times \frac{W}{f_c} \times D}$ for $x^{1:T}$, where $f_c$ is a downsample factor and $D$ is the number of hidden units. 
Since unique video signals express the frame-specific content of the input video clip, $E_u$ eliminates redundant information through temporal attention to obtain the unique latent features $z_u^{1:T}$.
Here, each unique latent feature $z_u^t \in R^{\frac{H}{f_u} \times \frac{W}{f_u} \times D}$ corresponds to the $t$-th video frame, where $f_u$ is another downsample factor.
\begin{align}
    z_c = E_c(x^{1:T}) \quad z_u^{1:T} = E_u(x^{1:T})
    \label{eq:zczu}
\end{align}
Notably, although the common latent feature $z_c$ and unique latent features $z_u^t$ initially exist as 2D features, we can expand them to 3D features by introducing an additional temporal dimension, which is set to 1. 
Then the resulting shapes for the common latent feature $z_c$ and each unique latent feature $z_u^t$ become $(1, H/fc,W/fc,D)$ and $(1, H/fu,W/fu,D)$, respectively.

To be specific, both $E_c$ and $E_u$ start with several convolution layers to obtain frame-wise video features with downsample factors $f_c$ and $f_u$, respectively.
Then $E_c$ concatenates the frame-wise video features along the channel dimension, and applies a convolution layer $g: R^{T \times D} \rightarrow R^D$ to diminish the temporal dimension and extract the common information from each spatial position, obtaining the common video feature.
On the other hand, $E_u$ concatenates the frame-wise video features along the temporal dimension and applies temporal self-attention to remove the redundant information, obtaining $T$ unique video features.
Subsequently, several residual layers are stacked to transform the common video feature into the common latent feature $z_c$, and to transform the unique video features into the unique latent features $z_u^{1:T}$, respectively.

\paragraph{Video Reconstruction}
For video reconstruction, the cascading merge module fuses the common latent feature with each unique latent feature, and the video decoder $D$ reconstructs each video frame of the input video clip based on the corresponding fused feature.
To be specific, given a common latent feature $z_c$ and $T$ unique latent features $z_u^{1:T}$, the cascading merge module separately fuses $z_c$ with each unique latent feature $z_u^t$ to produce the $t$-th fused video feature.
In particular, several downsample and upsample layers are applied to $z_c$ and $z_u^t$ respectively to obtain common and unique features with variable resolutions.
Then paired common and unique features with the same spatial resolution are concatenated and transformed into an intermediate feature.
As shown in Fig. \ref{fig:framework}, these intermediate features are then upsampled and summed in a cascading manner to incorporate multi-scale information, producing the $t$-th fused video feature.
Finally, the video decoder reconstructs the $t$-th video frame $\hat{x}^t$ based on the $t$-th fused video feature.
Specifically, several residual layers are stacked to transform fused video features and several convolution layers are employed to recover the spatial resolution and reconstruct video details.

\paragraph{Optimization}
Without label guidance, we train CU-VAE self-supervisedly with the target of video reconstruction.
In particular, the mean square error between the input and reconstructed video frames is used as the training target:
\begin{align}
    \mathcal{L}_{rec} = \frac{1}{T} \sum_{t=1}^{T} \lVert x^t - \hat{x}^t  \rVert_2
\end{align}
where $\lVert * \rVert_2$ denotes the calculation of the mean square error, $x^t$ is the $t$-th video frame of the input video clip and $\hat{x}^t$ is the $t$-th video frame of the reconstructed video clip.
Following \cite{ldm}, KL regularization is applied on both common and unique latent features to penalize them towards standard Gaussian distribution.

Following \cite{VQGAN}, we adopts a video discriminator $\mathcal{D}$ and an adversarial loss $\mathcal{L}_{adv}$ to improve the performance of video reconstruction:
\begin{align}
    \mathcal{L}_{adv} = log \mathcal{D}(x^{1:T}) + log(1 - \mathcal{D}(\hat{x}^{1:T}))
\end{align}
where $x^{1:T}$ denotes the input video clip and $\hat{x}^{1:T}$ denotes the reconstructed video clip.
LPIPS loss \cite{lpips} is used to stabilize the adversarial training process.

\subsection{Stage II: CU-LDM}
CU-LDM is composed of commonness and uniqueness diffusion streams and interpolated joint modules.
It models the common and unique latent features of the input video clip encoded by CU-VAE for generation following the diffusion theory.
A novel position embedding method is introduced to incorporate the absolute spatio-temporal positional information when modeling intermediate common and unique features.

\paragraph{Diffusion Streams}
The diffusion streams for modeling common and unique latent features conform similar forward and reverse diffusion processes.
We jointly call a common latent feature $z_c$ and a list of consecutive unique latent features $z^{1:T}_u$ as $z_0$ in the following part.
During the forward diffusion process, $z_0$ is corrupted by $S$ steps with the transition kernel:
\begin{equation}
    q_t(z_s|z_{s-1}) = \mathcal{N}(z_s;\sqrt{1-\beta_s}z_{s-1}, \beta_s \mathbf{I}) 
\end{equation}
where $s$ denotes the $s$-th step and $\beta_s$ is a hyper-parameter.
Obviously, the corrupted features $z_s$ from $s=1$ to $s=S$ construct a Markov chain:
\begin{equation}
    q(z_{1:S}|z_0) = \prod_{s=1}^S q(z_s|z_{s-1})
\end{equation}
In theory, when $S$ is large enough, the distribution of $z_S$ can be viewed as an isotropic Gaussian Distribution.
Furthermore, 
given $z_0$ and a set of hyper-parameters $\{\beta_s \in (0,1)\}_{s=1}^S$, the distribution of feature $z_s$ can be written as:
\begin{equation}
    q(z_s|z_0) = \mathcal{N}(z_s;\sqrt{\Bar{\alpha}_s} z_0, (1 - \Bar{\alpha}_s)\mathbf{I})
\end{equation}
where $\Bar{\alpha}_s = \prod_{i=1}^{s} \alpha_i$ and $\alpha_s = 1 - \beta_s$.
In other words, through simple reparameterization, we could obtain the corrupted feature at the $s$-th step directly by:
\begin{equation}
    z_s = \sqrt{\Bar{\alpha}_s} z_0 + \sqrt{1-\Bar{\alpha}_t} \epsilon_s
    \label{eq:zs}
\end{equation}
where $\epsilon_s$ is the noise feature randomly sampled from $\mathcal{N}(0, \mathbf{I})$, which has the same shape as $z_0$.

Based on the forward diffusion formulation, a diffusion model $p_{\theta}$ is trained to conduct the reverse diffusion process by simulating the distribution $q(z_{s-1}|z_s)$.
Thus a real feature $z_0$ can be obtained from a randomly sampled noise feature $\epsilon_S \sim \mathcal{N}(0, \mathbf{I})$ by performing reverse diffusion for $S$ steps.
However, the conditional probability $q(z_{s-1}|z_s)$ is unknown and untraceable, while given $z_0$ as condition, the conditional probability $q(z_{s-1}|z_s,z_0)$ is traceable and has known distribution formulation:
\begin{equation}
    q(z_{s-1}|z_s,z_0) = \mathcal{N}(z_{s-1}|\Tilde{\mu}_s(z_s,z_0), \Tilde{\beta}_s\mathbf{I})
\end{equation}
where $\Tilde{\mu}_s(z_s,z_0) = \frac{1}{\sqrt{\alpha_t}} (z_s - \frac{\beta_s}{\sqrt{1 - \Bar{\alpha}_s}} \epsilon_s)$, $\epsilon_s \sim \mathcal{N}(0, \mathbf{I})$ and $\Tilde{\beta}_s = \frac{1 - \Bar{\alpha}_{s-1}}{1 - \Bar{\alpha}_s} \beta_s$.
Considering that $\Tilde{\beta}_s$ is a deterministic constant, the only unknown term is $\epsilon_s$, namely the noise added in the $s$-th step.
Thus the diffusion model with parameter $\theta$ can be trained to predict the noise $\epsilon_s$ given the corrupted feature $z_s$ and the step-index $s$, obtaining the predicted noise $\epsilon_{\theta}(z_s, s)$.
Then the denoised feature $z_{s-1}$ can be obtained by:
\begin{equation}
    \begin{aligned}
        z_{s-1} &= \mathcal{N}(z_{s-1};\mu_{\theta}(z_s, s), \Tilde{\beta}_s\mathbf{I}) \\
        \mu_{\theta}(z_s, s) &= \frac{1}{\sqrt{\alpha_s}} 
        (z_s - \frac{\beta_s}{\sqrt{1 - \Bar{\alpha}_s}}\epsilon_{\theta}(z_s, s))
    \end{aligned}
\end{equation}
where $z_s$ can be easily obtained through Eq. (\ref{eq:zs}).

The commonness diffusion stream adopts the U-Net backbone improved by \cite{beat} to model common latent features like images.
Based on a similar U-Net model, the uniqueness diffusion stream appends an additional temporal attention layer after each spatial attention layer to capture temporal correlations between unique latent features.
As specified in stage II of Fig. \ref{fig:framework}, we integrate residual connections from the outputs of UNet Encoders (depicted in deep colors) to the outputs of UNet Decoders (depicted in shallow colors). 
These two latent features are concatenated along the channel dimension before being fed into the subsequent module, which is constructed using residual blocks.

\paragraph{Joint Modules}
As specified in Sec. III-A, the common latent feature captures shared video signals, including static scenes and object appearances, while the unique latent features focus on extracting frame-specific video content, such as object poses. 
Given that different objects exhibit distinct motion modes, it is crucial to ensure that the synthesized common and unique latent features contain compatible content. 
Based on this insight, we introduce joint modules to guarantee a consistent synthesis, thereby enhancing the realism of the generated videos. 
As depicted in Fig. \ref{fig:framework}, these joint modules are strategically interpolated across the commonness and uniqueness diffusion streams, facilitating an effective exchange of information between the common and unique latent features.
Given that the spatial resolution of common features ($\frac{HW}{f^2_c}$) surpasses that of unique features ($\frac{HW}{f^2_u}$), where $f_c<f_u$, we deliberately configure the commonness UNet encoder to include $log_2(\frac{f_u}{f_c})$ additional downsample convolution layers compared to the uniqueness UNet encoder. Accordingly, the commonness UNet decoder also includes $log_2(\frac{f_u}{f_c})$ additional upsample convolution layers compared to the uniqueness UNet decoder. In this way, we can obtain multiple intermediate common and unique features of the same resolution.

Given intermediate common and unique features with the same spatial resolution, the joint module first concatenates common and unique features along the temporal dimension, obtaining $z = [z_c:z_m^{1:T}] \in R^{(T+1) \times \frac{H}{f} \times \frac{W}{f} \times D}$, where $f$ is the downsample factor. 
Then query $z_q$, key $z_k$ and value $z_v$ are obtained by:
\begin{equation}
\begin{aligned}
z_q = zW_Q, z_k = zW_K, z_v = zW_V
\end{aligned}
\end{equation}
where $W_Q, W_K$ and $W_V$ are three learnable weight matrix $R^{D \times D'}$ and $D'$ is the number of hidden units.
To incorporate spatial absolute position information, height embeddings $he_q, he_k \in R^{\frac{H}{f} \times D'}$ and width embeddings $we_q, we_k \in R^{\frac{W}{f} \times D'}$ are adopted and shared for common and unique features.
To distinguish common features from unique features, specific temporal position embeddings $cte_q, cte_k \in R^{D'}$ are applied, obtaining common position embeddings $ce_q$ and $ce_k$:
\begin{align}
    ce_q^{h,w,d} = cte_q^{d} \times he_q^{h,d} \times we_q^{w,d} \\
    ce_k^{h,w,d} = cte_k^{d} \times he_k^{h,d} \times we_k^{w,d} 
\end{align}
Note that the temporal dimension of $ce_q$ and $ce_k$ is 1, which is skipped for concise.
And to capture the timing relationship of $T$ consecutive unique features, temporal absolute position embeddings $ute_q, ute_k \in R^{T \times D'}$ are employed, constructing unique position embeddings $ue_q$ and $ue_k$:
\begin{align}
    ue_q^{t,h,w,d} = ute_q^{t,d} \times he_q^{h,d} \times we_q^{w,d} \\
    ue_k^{t,h,w,d} = ute_k^{t,d} \times he_k^{h,d} \times we_k^{w,d} 
\end{align}
Notably, these embeddings are learnable and are optimized with the entire model.
The common and unique position embeddings are then concatenated along the temporal dimension and added to the query and key features:
\begin{align}
    z'_q = z_q + [ce_q:re_q], \quad z'_k = z_k + [ce_k:re_k]
\end{align}
where $[*:*]$ denotes the operation of concatenation.
Then the query $z_q'$, key $z_k'$ and value $z_v$ are spatially divided into several blocks to reduce computation complexity.
Each block has shape $(T+1)\times \frac{H}{fw} \times \frac{W}{fw} \times D'$, where $w$ is a hyper-parameter.
Finally, temporal-spatial attention is calculated based on matched query, key and value blocks.

Given that shapes of image features are predefined and typically maintain unchanged, our positional embedding incorporates absolute position information for three key reasons. 
First, when common and unique features are concatenated along the temporal dimension for self-attention calculation, the introduction of temporal embeddings for absolute temporal positions facilitates the distinction between common and unique features. 
Second, by defining spatial embeddings for absolute spatial positions and sharing these embeddings for both features, constraints are naturally imposed to enhance compatibility in the synthesis of common and unique features, prompting a harmonious integration. 
Third, the use of absolute positional embeddings ensures that block-wise attention is cognizant of both the global absolute position of each block and the local relative position within block features, thereby improving the spatial consistency of video generation.


\paragraph{Training Objective} 
Given common and unique latent features $z_c$ and $z_u^{1:T}$ and randomly sampled step-index $s \in \{1,2,...,T\}$, corrupted common and unique features can be obtained directly by Eq. (\ref{eq:zs}).
Following \cite{ddpm}, CU-LDM is trained to predict the added common noise $\hat{\epsilon}^c_s$ and unique noise $\hat{\epsilon}^u_s$ with an unweighted loss function:
\begin{align}
    \mathcal{L} := \mathbb{E}_{z_c, z^{1:T}_u, \epsilon^c, \epsilon^u, s}  \big [ \lVert \epsilon^c_s - \hat{\epsilon}^c_s \rVert_2 + \lVert \epsilon^u_s - \hat{\epsilon}^u_s \rVert_2 \big ]
\end{align}
where $\epsilon^c_s\sim\mathcal{N}(0,\mathbf{I})$ and $\epsilon^u_s\sim\mathcal{N}(0,\mathbf{I})$ are noise features that corrupt common and unique latent features at the $s$-th step.
\begin{table*}[t]
\centering
\caption{
    Quantitative comparison with state-of-the-art methods for unconditional video generation.
}
\label{tab:gen}
\begin{minipage}[htp]{0.35\linewidth}
    \begin{minipage}[htp]{1\linewidth}
        \centering
        \subcaptionbox{\raisebox{0.5\height}{FaceForensics $256^2$}}{
        \vspace{3mm}
        \scalebox{1}{
        \begin{tabularx}{1\linewidth}{ p{0.6\linewidth} | X<{\centering} }
            \toprule
            Model                               &  FVD$\downarrow$ \\
            \midrule
            VideoGPT \cite{videogpt}            & 185.9     \\
            MoCoGAN \cite{mocogan}              & 124.7     \\
            ND                                  & 117.6     \\
            MoCoGAN-HD \cite{mocogan-hd}        & 111.8     \\
            DIGAN \cite{digan}                  & 62.5      \\
            \midrule
            \rowcolor{gray!30}
            COMUNI (ours)                       & 55.2      \\
            \bottomrule
        \end{tabularx}}}
    \end{minipage}
    \begin{minipage}[htp]{1\linewidth}
        \centering
        \subcaption{\raisebox{0.5\height}{UCF101 $256^2$}}
        \scalebox{1}{
        \begin{tabularx}{1\linewidth}{ p{0.6\linewidth} | X<{\centering} }
            \toprule
            Model                               & FVD$\downarrow$ \\
            \midrule
            MoCoGAN \cite{mocogan}              & 3679.0     \\
            MoCoGAN-HD \cite{mocogan-hd}        & 2606.5     \\
            DIGAN \cite{digan}                  & 2293.7      \\
            StyleGAN-V \cite{stylegan-v}        & 1773.4     \\
            ND                                  & 1343.9   \\
            MOSO \cite{moso}                    & 1202.6    \\
            \midrule
            \rowcolor{gray!30}
            COMUNI (ours)                       & 773.7      \\
            \bottomrule
        \end{tabularx}}
    \end{minipage}
\end{minipage}
\hspace{6mm}
\begin{minipage}[htp]{0.495\linewidth}
    \centering
    \subcaption{\raisebox{0.5\height}{UCF101 $128^2$}}
    \scalebox{1}{
    \begin{tabularx}{1\linewidth}{ p{0.4\linewidth} | X<{\centering} X<{\centering} X<{\centering} }
        \toprule
        Model       & Class &   IS$\uparrow$    & FVD$\downarrow$   \\
        \midrule
        TGAN \cite{tgan}        &\xmark &   11.85   &   -   \\
        TGAN \cite{tgan}       &\cmark &   15.83   &   -   \\
        MoCoGAN \cite{mocogan}          &\cmark &   12.42(${\pm.07}$)   &   -   \\
        LDVD-GAN \cite{ldvd}            &\xmark &   22.91(${\pm}.19$)   &   -   \\
        VideoGPT \cite{videogpt}        &\xmark &   24.69(${\pm}.30$)   &   -   \\
        TGANv2 \cite{tganv2}            &\cmark &   28.87(${\pm}.67$)   &   1209(${\pm}28$)    \\
        DVD-GAN \cite{dvd}              &\cmark &   27.38(${\pm}.53$)   &   -   \\
        MoCoGAN-HD \cite{mocogan-hd}    &\xmark &   32.36               &   838 \\
        DIGAN \cite{digan}              &\xmark &   29.71(${\pm}.53$)   &   655(${\pm}$22) \\
        DIGAN* \cite{digan}             &\xmark &   32.70(${\pm}.35$)   &   577(${\pm}$21) \\
        CCVS+Real frame \cite{ccvs}     &\xmark & 41.37(${\pm}.39$)     &   389(${\pm}$14) \\
        CCVS+StyleGAN \cite{ccvs}       &\xmark &   24.47(${\pm}.13$)   &   386(${\pm}$15) \\
        StyleGAN-V \cite{stylegan-v}    &\xmark &   23.94(${\pm}.73$)   &   -   \\
        CogVideo \cite{cogvideo}        &\cmark &   50.46               &   626 \\
        VDM \cite{vdm}                  &\xmark & 57.00(${\pm}.62$)     &   -   \\
        TATS \cite{tats}                &\xmark &  57.63(${\pm}.24$)    &   420(${\pm}$18) \\
        MMVG \cite{mmvg}                &\xmark &   58.3    & 395   \\
        VIDM \cite{vidm}                &\xmark &   64.2                &   263   \\
        VideoFusion \cite{videofusion}  &\xmark &   72.2                &   220   \\
        \midrule
        \rowcolor{gray!30} 
        COMUNI (ours)                   &\xmark &  73.1(${\pm}.15$)    &   210(${\pm}$8) \\
        \bottomrule
    \end{tabularx}}
\end{minipage}
\end{table*}

\begin{table}[t]
    \centering
    \caption{
    Comparison of sampling time/memory using different methods for generating multiple video frames with resolution of 256$^2$, batch size of 1, 100 diffusion steps, and comparable GPU memory on a v100 GPU.
    $F$ represents the number of video frames.
    }
    \label{tab:speed}
    \scalebox{0.8}{
        \begin{tabularx}{1.2\linewidth}{ p{0.15\linewidth} |  X<{\centering} X<{\centering} X<{\centering} X<{\centering} X<{\centering}}
        \toprule
        Method              & VIDM 
                            & VDM
                            & LVDM
                            & ModelScope
                            & COMUNI       \\
        & \cite{vidm}&\cite{vdm}&\cite{lvdm}&\cite{videofusion} &(ours) \\
        \midrule
        w/ VAE  & \xmark & \xmark & \cmark & \cmark & \cmark   \\
        \midrule
        $F=16$          &192s/20G &125s/11G &113s/9G  &39s/6G &\textbf{16s}/12G      \\
        $F=32$          &375s/20G &234s/11G &212s/13G &64s/8G &\textbf{46s}/12G    \\
        $F=64$          &771s/20G &329s/11G &432s/20G &123s/12G &\textbf{110s}/12G    \\
        \bottomrule
        \end{tabularx}
    }
\end{table}


\section{Experiments}
In this section, a series of experiments are conducted to compare our proposed methodology against the state-of-the-art models. 
These experiments demonstrate the efficacy of decomposing common and unique video signals for video generation, as well as the efficiency of separating video decomposition from video generation.

\paragraph{Datasets and Evaluations}
We show results on FaceForensics \cite{ffs} and UCF-101 \cite{ucf} datasets for unconditional video generation with resolution $128^2$ and $256^2$.
Following \cite{stylegan-v}, we use train split for FaceForensics and train+test splits for UCF101 and preprocess videos in FaceForensics to crop human faces.
We adopt the Fr$\Acute{e}$chet Video Distance (FVD) \cite{FVD} and Inception Score (IS) metric implemented by \cite{stylegan-v} based on the C3D model to evaluate the realism of generated videos following previous works \cite{tats,digan,mocogan} except otherwise specified.
In particular, 2048 video samples are generated to calculate FVD with reference to 2048 randomly selected real videos, and 10000 video samples are synthesized to calculate IS.
SSIM \cite{ssim}, PSNR \cite{psnr} and LPIPS \cite{lpips} are employed to evaluate the performance of VAE models by calculating scores between each input video clip and corresponding reconstructed video clip by frame.

\paragraph{Training Details}
For a fair comparison with previous works \cite{stylegan-v,digan}, the COMUNI model is trained using 16-frame video clips. 
The downsample factors, i.e. $f_c$ and $f_u$, are set as 4 and 16 respectively except otherwise specified.
The number of channels for both common and unique latent features is fixed as 3.
Importantly, as the spatial resolution of each common latent feature is four times larger than that of each unique latent feature, the number of elements in all unique latent features remains equal to that of the common latent features. 
For example, the dimensions of the unique and common latent features are $16 \times 8 \times 8 \times 3$ and $32 \times 32 \times 3$, respectively, for input video clips with a spatial resolution of $128^2$, where both the unique and common latent features contain 3072 elements in total.

When training CU-VAE, the adversarial training loss $\mathcal{L}_{adv}$ is adopted after 150K iterations.
The patch-wise discriminator \cite{VQGAN} is adopted and the loss weight of $\mathcal{L}_{adv}$ is fixed to be 0.1.
The loss weight for the LPIPS loss \cite{lpips} is 1.
Following \cite{ldm}, we adopt a loosen loss of KL regularization with loss weight 1e-6.
For all VAE-based models, we test the reconstruction performance of last 4 checkpoints and employ the checkpoint with the smallest FVD score.

When training CU-LDM, we calculate the mean square error loss between predicted noises and real noises of common and unique features respectively.
Following \cite{sde}, we adopt continuous timesteps during training.
With the help of the fast sampling strategy dpm-solver \cite{dpm}, we randomly sample 2048 generated videos of each checkpoint and select the checkpoint with the smallest FVD to be the final model.
Additional 2048 videos are generated through the traditional DDPM sampler with 1000 timesteps \cite{ddpm} by the final model to obtain the final FVD score.
The detailed settings of other hyper-parameters are given in the appendix.

\subsection{Quantitative Comparison with State-of-the-Art Models}
\paragraph{Generation Quality}
We compare COMUNI with state-of-the-art methods on three benchmarks: FaceForensics $256^2$, UCF-101 $256^2$ and UCF-101 $128^2$.
In order to assess the effectiveness of decomposing common and unique video signals, we also present the results of a baseline non-decomposition video generation method, referred to as ND. 
Similar to the COMUNI approach, ND employs a two-stage framework consisting of a video VAE (ND-VAE) for encoding video clips into latent features, and a one-stream latent diffusion model (ND-LDM) for generating videos based on the latent features. 
ND-VAE encodes input video clips into frame-wise latent features and is trained using the same loss functions as CU-VAE. 
To capture temporal correlations between frame-wise latent features, ND-LDM is built on the U-Net backbone, which has been enhanced by \cite{beat}, and includes a temporal attention layer after each spatial attention layer. 
To ensure a fair comparison, the parameters of ND and COMUNI are set to be comparable, and the training hyperparameters (e.g., batch size, learning rate, number of hidden units, and training iterations) are identical. 
As reported in Table \ref{tab:gen}, 
On the FaceForensics dataset, COMUNI outperforms previous best work by 7.3 FVD.
On the UCF-101 $256^2$ benchmark, MOSO \cite{moso} obtains the previous best results for unconditional video generation.
Compared with MOSO, our COMUNI further reduces the FVD score by 428.9.
On the UCF-101 $128^2$ benchmark, our COMUNI outperforms prior best model VIDM \cite{vidm} by 42 FVD and 1.6 IS.

\paragraph{Generation Efficiency}
To enhance video generation efficiency, we decompose and encode video signals into common and unique features, thereby reducing redundant modeling of common video signals. 
We compare the generation efficiency of our proposed method with prior models and report the results in Table \ref{tab:speed}.
It can be seen that our COMUNI surpasses one-stage methods like VIDM and VDM, as well as two-stage methods like LVDM and ModelScope, in terms of generation efficiency.
This superiority can be attributed to two key factors. 
Firstly, we represent videos with low-dimensional features through Video Auto-Encoder (VAE), thus reducing the computation complexity compared to one-stage models.
Secondly, our VAE decomposes common and unique video signals during the encoding process. In this way, we alleviate video generative models from modeling redundant information, thus further enhancing the generation efficiency and achieving superior efficiency compared to other two-stage models.


\begin{figure}[t]
    \centering
    \includegraphics[width=1.0\linewidth]{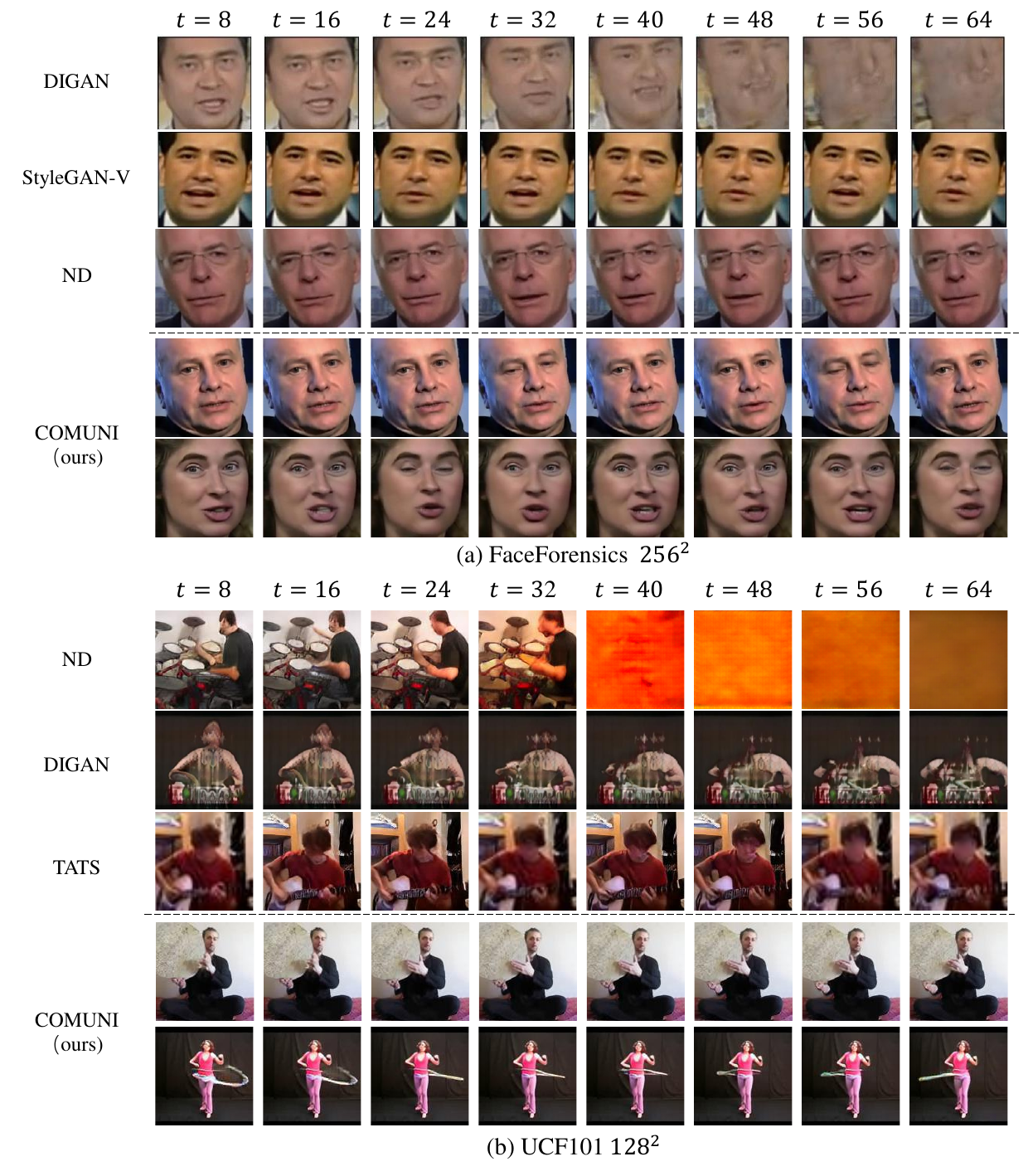}
    \caption{Qualitative comparison with other models for video generation on FaceForensics $256^2$ and UCF-101 $128^2$.}
    \label{fig:generation}
\end{figure}

\begin{figure}[t]
    \centering
    \includegraphics[width=1.0\linewidth]{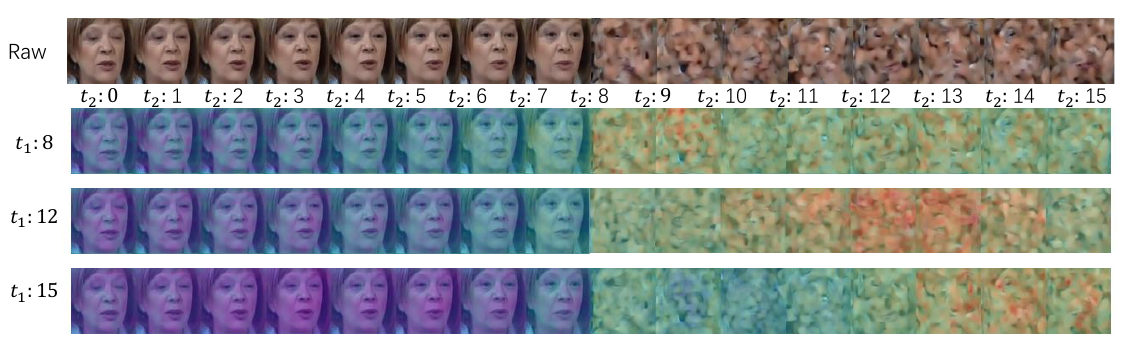}
    \caption{
    Visualization of the temporal attention map in CU-LDM for conditional generation of the subsequent 8 unique features based on 8 synthesized unique features.
    We employ $t_1$ to denote the row temporal index and $t_2$ to denote the column temporal index of the map.
    }
    \label{fig:long_attn_visualize}
\end{figure}
\begin{figure}[t]
    \centering
    \includegraphics[width=1.0\linewidth]{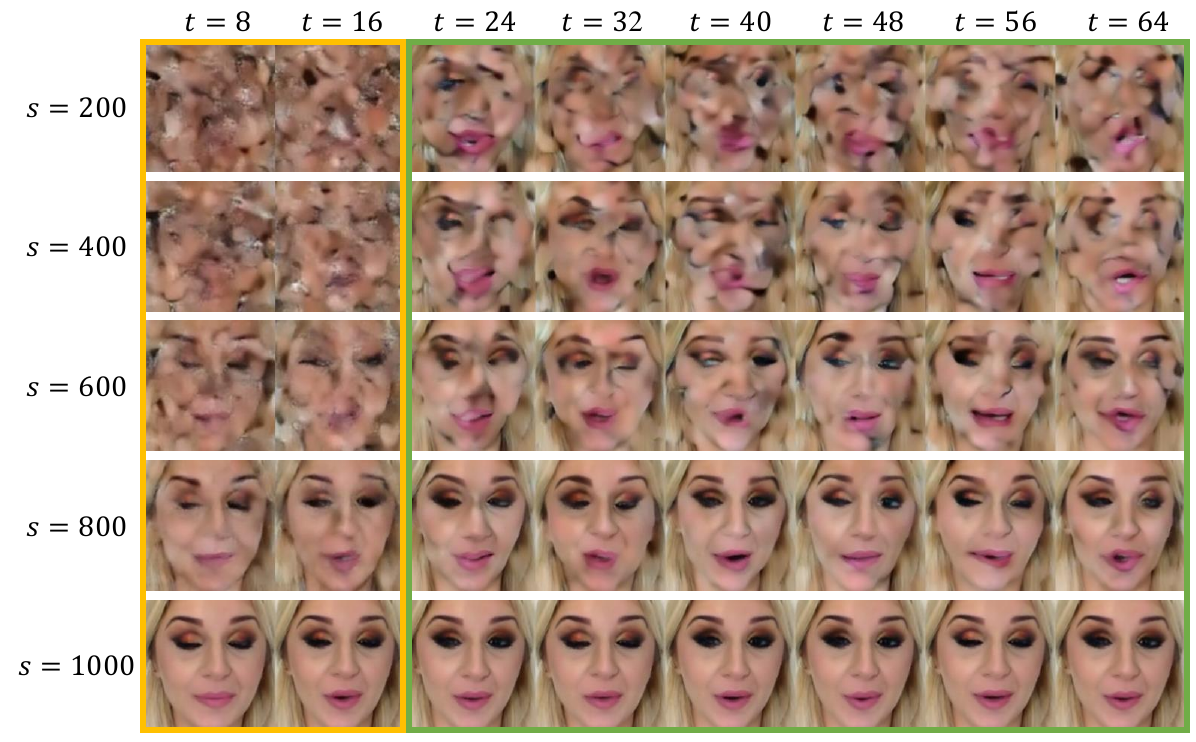}
    \caption{
    Visualization of a synthesized long video in distinct sampling steps.
    Video frames in the yellow rectangle are unconditionally generated as an initial video clip, 
    and video frames in the green rectangle are conditionally produced using the iterative generation method.
    }
    \label{fig:long_visualize}
\end{figure}
\subsection{Qualitative Results for Long Video Generation}
For qualitative evaluation, we conduct comparison between COMUNI and state-of-the-art methods on two benchmark datasets, namely FaceForensics $256^2$ and UCF-101 $128^2$, are shown in Fig. \ref{fig:generation}.
When synthesizing videos with a length of 64 frames, COMUNI follows an iterative generation process.
It first generates common and unique latent features for an initial 16-frame video clip.
Then, the common latent feature and the last 8 unique latent features (referred to as conditional features) are held constant and fed into CU-LDM to produce the next 8 unique latent features (referred to as target features).
Despite the iterative generation process introducing varied noise levels between conditional and target features, such inconsistency brings negligible influence to the synthesis of target features for two reasons:
1) CU-LDM processes each unique feature individually except temporal attention layers and joint modules.
2) The attention mechanism in both temporal attention layers and joint modules can adaptively capture valuable information from conditional features and assign more weights to target features, as depicted in Fig. \ref{fig:long_attn_visualize}.
In the end, the denoising process of target features (with condition features) becomes consistent with that of the initial video clip (without condition features) after 800 steps as shown in Fig. \ref{fig:long_visualize}.
This iterative generation method allows us to obtain videos of varying lengths with consistent video content since the common latent feature is kept constant.
Similarly, ND first generated 16 consecutive latent features of a 16-frame video, then used the last 8 latent features to generate the subsequent 8 video frames to obtain a long video.

On the FaceForensics dataset, DIGAN \cite{digan} fails to generate realistic human faces for the last few dozens frames of the generated video.
Although StyleGAN-V \cite{stylegan-v} and ND can generate long videos with clear motion, their generated videos tend to have slightly blurred details.
In contrast, our COMUNI is able to generate videos with the clear human appearance and significant movements.
On the UCF-101 benchmark, both ND and DIGAN fail to generate coherent video content and produce meaningful frames for long videos.
This may be caused by their lack of mechanism to record global video content and maintain it.
TATS \cite{tats} generates videos with blurred human appearances and fails to maintain object details for the last few dozens frames of videos.
StyleGAN-V \cite{stylegan-v} and VIDM \cite{vidm} can not well synthesize human details such as hands.
Compared with the previous methods, our COMUNI generates videos with superior content consistency and the most distinct appearances.
This is due to two aspects.
Firstly, video consistency is easy to obtain for COMUNI since it can keep the common latent feature unchanged, which records the common video content that shared by all video frames, when generating long videos (i.e. generating unique latent features for frames of long videos).
Secondly, when encoding common video signals to latent features, we typically adopt a relatively small commonness downsample factor. 
This approach allows for a reduction in the loss of common video information, resulting in more reserved video details and thus more distinct object appearances. 
More video samples generated by our proposed COMUNI are presented in: 
\url{https://anonymouss765.github.io/COMUNI}.

\begin{figure}[t]
    \centering
    \includegraphics[width=1.0\linewidth]{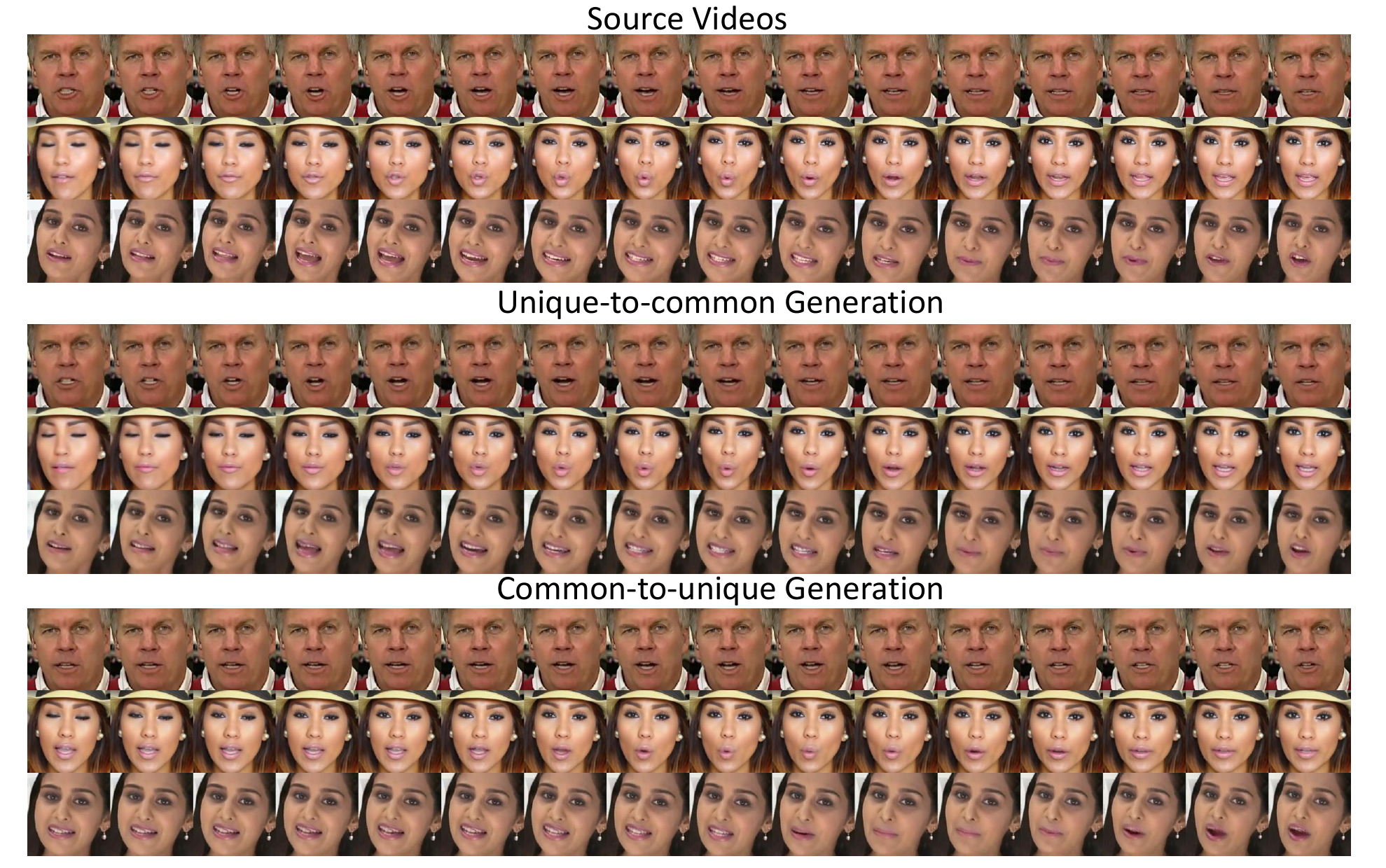}
    \caption{
    Qualitative results of common-to-unique and unique-to-common generation. 
    Source videos are used to obtain conditional common and unique features.
    }
    \label{fig:conditional_gen}
\end{figure}

\begin{figure}[t]
    \centering
    \includegraphics[width=1.0\linewidth]{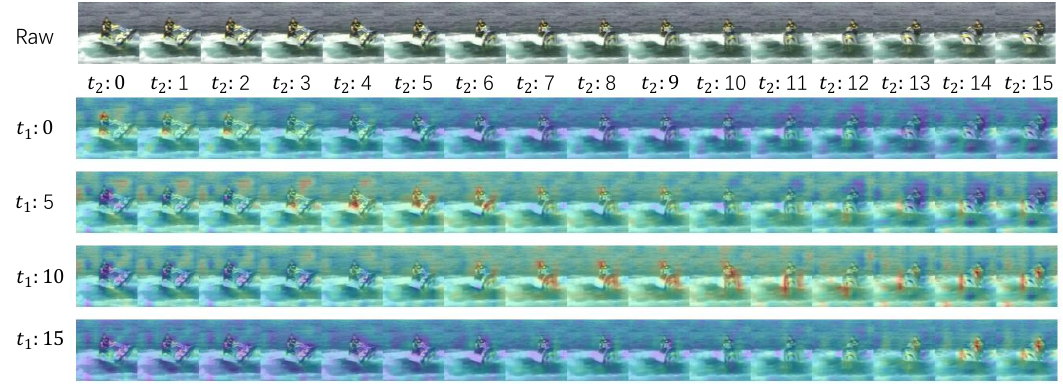}
    \caption{
    Visualization of the temporal attention map in the uniqueness video encoder of CU-VAE.
    We employ $t_1$ to denote the row temporal index and $t_2$ to denote the column temporal index of the map.
    }
    \label{fig:temporal_attn}
\end{figure}

\subsection{Qualitative Results for Conditional Video Generation}
To take a deep insight of CU-LDM and explore whether the common and unique features contain overlapped information, we employ CU-LDM for conditional generation—specifically, common-to-unique and unique-to-common generation—using the FaceForensics dataset. 
The former synthesizes common features based on unique features, and the later synthesizes unique features based on common features.
Notably, if the common feature of a given video clip does not incorporate unique information, then the synthesized video clip should have the same human appearance with different expressions. If unique features of a given video clip do not incorporate common information, then the synthesized video clip should have the same human expressions with different appearances. If the common feature of a given video clip does incorporate unique information or the unique features do incorporate common information, then the synthesized video clip should be the same as the given video clip. 

The results are presented in Fig. \ref{fig:conditional_gen}. For the common-to-unique generation, the model produces video clips with consistent human appearance but varying expressions, indicating that common features do not incorporate unique information. 
It is reasonable since the common feature is shared between video frames when decoding, constraining the common feature to integrate common information.
For the unique-to-common generation, the synthesized video clips are the same as input video clips, demonstrating that unique features contain redundant common information.
However, given the smaller spatial resolution of unique features ($\frac{1}{4}$ of that of common features), such redundancy is acceptable for video generation.

\subsection{Visualization of Temporal Attention}
As specified in Sec. \ref{sec:cu-vae}, the uniqueness video encoder employs temporal attention to extract unique video signals for each video frame. To provide deeper insights into the functioning of temporal attention, we visualize the temporal attention maps in Fig. \ref{fig:temporal_attn}. It can be seen that the $t_1$-th video frame assigns significance to distinct content within adjacent frames (i.e. object motions), and to common content within remote frames (i.e. video scenes). It is reasonable since tracking movements across adjacent frames enables the capture of unique signals, while comparing scenes with remote frames facilitates the identification and removal of common video signals.

\subsection{Qualitative Results on a Large-scale Dataset}
\begin{figure}[t]
    \centering
    \includegraphics[width=1.0\linewidth]{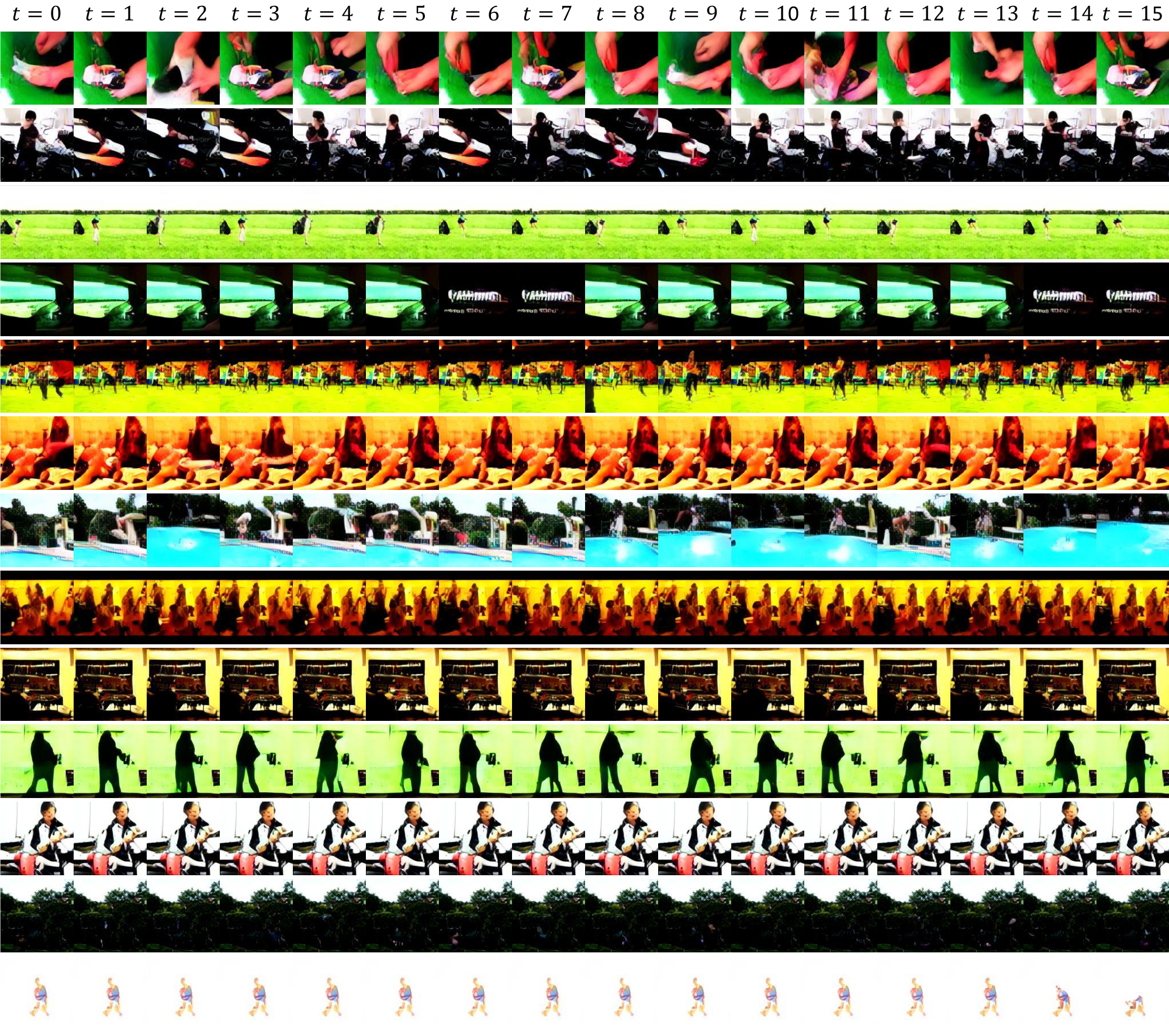}
    \caption{
    Video samples on the Kinetics-400 dataset \cite{kinetics400}.
    }
    \label{fig:kinetics-400}
\end{figure}
To further evaluate the generation capability of our proposed method, we train COMUNI on the Kinetics-400 dataset \cite{kinetics400}.
In particular, we extract 4 frames per second and train COMUNI to synthesize 16 frames in each sample.
The synthesized videos are presented in Fig. \ref{fig:kinetics-400}. 
It can be seen that our proposed method performs well on the large-scale dataset.

\subsection{Ablation Study}

\paragraph{Ablate Video Decomposition}
\begin{table*}[ht]
    \centering
    \caption{
    Ablation study on video decomposition on FaceForensics and UCF101 datasets for video reconstruction.
    ND denotes the non-decomposition method.
    VR denotes video reconstruction.
    VG denotes video generation.
    }
    \label{tab:ablate_decomposition}
    \scalebox{1}{
    \begin{tabularx}{1\linewidth}{ p{0.1\linewidth} | X<{\centering} X<{\centering} X<{\centering} X<{\centering} | X<{\centering} | X<{\centering} X<{\centering} X<{\centering} X<{\centering} | X<{\centering} }
    \toprule
    \multirow{3}{*}{Model}  &   \multicolumn{5}{c|}{FaceForensics}      &   \multicolumn{5}{c}{UCF101} \\ 
    \cline{2-11}
                            &   \multicolumn{4}{c|}{VR}  & VG           &   \multicolumn{4}{c|}{VR}  & VG \\
    \cline{2-11}
    & PSNR$\uparrow$    & SSIM$\uparrow$    & LPIPS$\downarrow$ & rFVD$\downarrow$ & FVD$\downarrow$ 
    & PSNR$\uparrow$    & SSIM$\uparrow$    & LPIPS$\downarrow$ & rFVD$\downarrow$ & FVD$\downarrow$ \\
    \midrule
    ND$_{128}$  & \cellcolor{gray!30} 36.0  & \cellcolor{gray!30}  96.6  & 0.037 & 11.6 &  84.9 
                    & 27.8  & 75.4  & 0.097 & 74.7  & 620.8  \\ 
    \rowcolor{gray!30}
\cellcolor{white} COMUNI$_{128}$  & \cellcolor{white} 34.7  & \cellcolor{white} 96.1  & 0.011 & 5.1 &  45.5 
                    & 35.1  & 92.5  & 0.031 & 27.9  &  221.0   \\
    \midrule
    ND$_{256}$  & 30.9  & 88.0  & 0.116 & 30.2  &  117.6 
                    & 26.3  & 68.2  & 0.194 & 153.7 &  1343.9 \\
    \rowcolor{gray!30}
\cellcolor{white} COMUNI$_{256}$  & 34.4  & 94.5  & 0.042 & 24.8  &  55.2 
                    & 32.5  & 87.9  & 0.079 & 55.1  &  773.7  \\
    \bottomrule
    \end{tabularx}}
\end{table*}
In order to assess the effectiveness of decomposing common and unique video signals, we devised a baseline non-decomposition video generation method, referred to as ND. 
As reported in Table \ref{tab:ablate_decomposition}, We compare COMUNI with ND for video reconstruction and video generation on the FaceForensics and UCF-101 datasets with resolutions of $128^2$ and $256^2$.
When reconstructing videos, COMUNI utilizes CU-VAE to decompose video signals and encode them into latent features, while ND utilizes ND-VAE to encode input video clips into frame-wise latent features. 
Both CU-VAE and ND-VAE decode latent features into video clips and are trained self-supervised using the same loss functions. 
High-quality reconstructed videos are important, as they determine the upper-bound performance of video generation. 
However, CU-VAE outperforms ND-VAE on the video reconstruction task on almost all metrics. 
This may be due to two factors. 
Firstly, when the input video has a spatial resolution of $128^2$, both the number of feature elements in a common feature $z_c \in R^{\frac{H}{f_c}\times\frac{W}{f_c}\times D}$ and all the 16 unique features $z_u^t \in R^{\frac{H}{f_u}\times\frac{W}{f_u}\times D}$, where $t={1,2,...,16}$, amount to 3072, given $f_c=4$, $f_u=16$, and $D=3$ as specified in the training details.
Considering that the element number of video features in ND-VAE is equivalent to the total number of elements in unique features in CU-VAE, the total number of all feature elements in CU-VAE is twice that of ND-VAE. 
This characteristic allows CU-VAE to retain more video information compared to ND-VAE.
Secondly, by decomposing common and unique video signals, redundant information is efficiently recorded in the common latent feature through CU-VAE, while the frame-wise unique latent features focus on recording non-overlapping information. 
In contrast, the frame-wise latent features in ND-VAE must record all types of information, resulting in overlapped information between frame-wise latent features and a loss of recording capacity.

In terms of video generation, it is notable that ND has a much higher FVD score than COMUNI, despite having fewer latent features than COMUNI. 
This phenomenon may be attributed to two aspects. 
Firstly, ND-LDM relies on temporal attention to obtain consistent video content when synthesizing frame-wise latent features, while content consistency is straightforward for videos generated by COMUNI since all video frames share the same content latent feature.
This observation highlights the efficacy of decomposing common and unique video signals for video generation.
Secondly, the joint modules of COMUNI incorporate absolute position information through our proposed position embedding method, which is effective in maintaining spatial consistency and motion coherence in the generated videos.

\paragraph{Ablate the Joint Module}
\begin{table}[t]
    \centering
    \caption{
    Ablation study on the joint module and the position embedding method.
    JM denotes the joint module.
    PE denotes the position embedding method proposed for the joint module.
    KV and QK denote adding position embeddings to key and value features or query and key features respectively.
    }
    \scalebox{1}{
        \begin{tabularx}{1\linewidth}{ p{0.2\linewidth}<{\centering} | X<{\centering} X<{\centering} X<{\centering} | p{0.2\linewidth}<{\centering} }
            \toprule
            Method    & $f_{u}$   & JM  & PE & FVD \\
            \midrule
            ND      &   16       & \xmark  & \xmark    &  117.6       \\
            \midrule
            \multirow{5}{*}{COMUNI}      &   16       & \xmark  & \xmark    &  82.5 \\
                  &   16       & \cmark  & \xmark    &  80.8     \\
                  &   16       & \cmark  & QK        &  65.8     \\
                  &   16 \cellcolor{gray!30}      &\cellcolor{gray!30} \cmark  &\cellcolor{gray!30} KV        &\cellcolor{gray!30}  55.2     \\
                  &   8        & \cmark  & KV        &  133.1     \\
            \bottomrule
    \end{tabularx}}
    \label{tab:ablat-ldm}
\end{table}

We ablate CU-LDM on the joint module, the position embedding method, and the uniqueness downsample factor $f_u$.
The results are reported in Table \ref{tab:ablat-ldm}.
To keep the temporal and spatial consistency of the common and unique diffusion streams, joint modules adopt the block-wise temporal-spatial attention mechanism to capture both temporal and spatial relationships.
By replacing the joint module with simple temporal attention, the FVD score further increases to 82.5, which demonstrates the effectiveness of the joint module.

Considering that query and key features are used to calculate attention weights, which finally multiply value features to obtain attention results, we compare two situations of adding the position embeddings: 
1) adding the position embeddings to key and value features, which incorporates position information into both attention weights and values;
2) adding the position embeddings to query and key features, which only incorporates position information into attention weights.
As shown in Table \ref{tab:ablat-ldm}, compared to the first situation, the second situation increases FVD by 10.6, which demonstrates the necessity of incorporating position information into value features.
When we do not employ the position embeddings, the temporal motion coherence can not be modeled and the attention calculation can not distinguish between common and unique video features.
Moreover, the block division operation may disturb spatial relationships, thus leading to a significant increase in the FVD score to 80.8.

Without decomposing common and unique video signals, the ND generates videos with FVD 117.6, which is much higher than that of CU-LDM, demonstrating the effectiveness of decomposing common and unique video signals.
When we decrease the uniqueness downsample factor $f_u$, the spatial resolution of unique latent features increases.
Thus the modeling of unique latent features becomes more difficult, leading to the FVD increasing to 133.1.

\paragraph{Ablate on the $\mathcal{L}_{GAN}$ Loss}
We conduct an ablation study on CU-VAE to explore the necessity of the adversarial training loss $\mathcal{L}_{GAN}$.
The results are reported in Table \ref{tab:ablate-vae}.
After removing the adversarial loss, the reconstruction FVD score increases 14.7 and 18.9 on UCF-101 $128^2$ and $256^2$, and 0.3 and 10.1 on FaceForensics $128^2$ and $256^2$.
By employing adversarial training, the discriminator helps figure out differences between the input and reconstructed video frames, thus CU-VAE could improve its reconstruction performance by eliminating the differences and obtain more realistic reconstructed videos.
\begin{table}[t]
    \centering
    \caption{Ablate on the $\mathcal{L}_{GAN}$ loss when training CU-VAE.}
    \scalebox{1}{
        \begin{tabularx}{1\linewidth}{ X<{\centering} | X<{\centering} X<{\centering}  | p{0.3\linewidth}<{\centering} }
            \toprule
            Datasets        &   Resolution  &   $\mathcal{L}_{GAN}$   & rFVD $\downarrow$  \\
            \midrule
            \multirow{4}{*}{UCF101}         & \multirow{2}{*}{$128^2$}  & \cmark  & \textbf{28.2} \\
                                            &                           & \xmark  & 42.9 \\
            \cline{2-4}
                                            & \multirow{2}{*}{$256^2$}  & \cmark  & \textbf{55.1}  \\
                                            &                           & \xmark  & 74.2  \\
            \midrule
            \multirow{4}{*}{FaceForensics}  & \multirow{2}{*}{$128^2$}  & \cmark  & \textbf{4.8}   \\
                                            &                           & \xmark  & 5.1   \\
            \cline{2-4}
                                            & \multirow{2}{*}{$256^2$}  & \cmark  & \textbf{9.2}  \\
                                            &                           & \xmark  & 10.3  \\
            \bottomrule 
        \end{tabularx} }
    \label{tab:ablate-vae}
\end{table}

\section{Limitation}
\begin{figure}[t]
    \centering
    \includegraphics[width=0.9\linewidth]{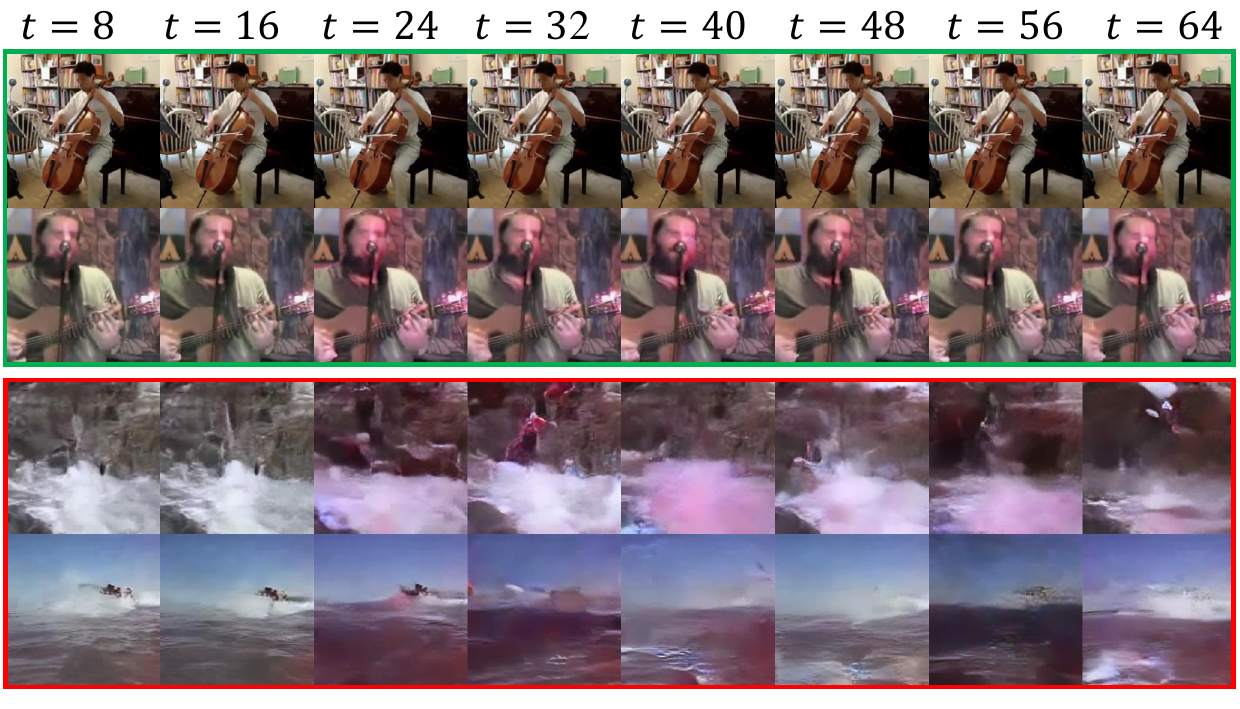}
    \caption{Videos in the green rectangle are successful samples with much commonness. 
    Videos in the red rectangle are failed samples with rapidly changed scenes.
    }
    \label{fig:limitation}
\end{figure}

Despite our proposed method could generate realistic videos with distinct objects, we find it suffers from three major limitations.
Firstly, it is difficult to generate videos with fast-changing backgrounds as shown in Fig. \ref{fig:limitation}.
Compared to the commonness downsample factor $f_c$, we adopt a much larger uniqueness downsample factor $f_u$, e.g. 16 vs 4.
When real videos contain little commonness, e.g. videos with fast-changing scenes, the common latent features count for little.
Thus the unique latent features have to encode much more useful information, and content consistency can not be ensured during the generation process due to the lack of valid common video signals.
Secondly, when generating long videos, repetitive movements are observed in the video samples. 
This phenomenon arises from our fixed common features throughout the generation of long videos, imposing overly strong constraints on the subsequent generation of unique features and hindering the production of diverse motions.
Thirdly, we find our unique latent features encompass information that related to common features.

\section{Conclusion}
In this paper, we propose a novel two-stage framework named COMUNI, which decomposes common and unique video signals for the video generation task.
In the first stage, CU-VAE is proposed to decompose common and unique video signals by extracting them with two specific video encoders, obtaining common and unique latent features respectively.
A merge module is adopted to recompose video signals by fusing common and unique latent features and a video decoder is used to decode fused video features to reconstructed videos.
Thus CU-VAE can be trained in a self-supervised manner.
Then in the second stage, CU-LDM is employed to model common and unique latent features with a common diffusion stream, a unique diffusion stream and interpolated joint modules.
To distinguish common features from unique features and incorporate absolute position information, a novel position embedding method is used in each joint module by adding specific embeddings to key and value features when calculating attention.
Extensive experiments demonstrate the effectiveness and efficiency of our proposed method, and the importance of decomposing common and unique video signals.



\bibliographystyle{IEEEtran}
\bibliography{refer}

\newpage
 




\vfill

\end{document}


\title{ COMUNI: Decomposing Common and Unique Video Signals for Diffusion-based Video Generation }

\author{
\IEEEauthorblockN{
Mingzhen Sun\IEEEauthorrefmark{1,2},
Weining Wang\IEEEauthorrefmark{1},
Xinxin Zhu\IEEEauthorrefmark{1}, and
Jing Liu\IEEEauthorrefmark{1,2}} \\
\IEEEauthorblockA{\IEEEauthorrefmark{1}The Laboratory of Cognition and Decision Intelligence for Complex Systems, Institute of Automation, Chinese Academy of Sciences, Beijing, China} \\
\IEEEauthorblockA{\IEEEauthorrefmark{2}School of Artificial Intelligence, University of Chinese Academy of Sciences, Beijing, China} \\
\IEEEauthorblockA{sunmingzhen2020@ia.ac.cn, \{weining.wang, xinxin.zhu\}@nlpr.ia.ac.cn} \\
\IEEEauthorblockA{Corresponding Author: Jing Liu \quad Email: jliu@nlpr.ia.ac.cn}
}

\markboth{Journal of IEEE Transactions on Multimedia}%
{Shell \MakeLowercase{\textit{et al.}}: A Sample Article Using IEEEtran.cls for IEEE Journals}


\maketitle

{\appendices
\section*{Ablation on Sampling Strategies}

For an iterative generation, CU-LDM first samples 8 unique noise features from an isotropic Gaussian distribution and then predicts corresponding noises based on the conditional latent features, e.g. the common latent feature and the previous last 8 unique latent features.
In this section, we conduct an ablation study on the changes of the conditional latent features during the iterative generation process.
In particular, six sampling strategies are considered:
\begin{enumerate}
    \item All conditional latent features are changed in the same way as the input unique noise features.
    \item During each generation process, all conditional latent features are changed in the same way as the input noise features. While during the end of each iterative generation, the common latent feature is replaced by the starting one.
    \item The predicted noise feature of the common latent feature is abandoned and the common latent features are fixed during the entire process of iterative generation. While the conditional unique latent features are changed in the same way as the input noise features.
    \item The conditional common latent feature is changed in the same way as the input noise features, while the conditional unique latent features are fixed.
    \item The conditional common latent feature is changed in the same way as the input noise features during the generation process, but is replaced by the starting one at the end of each iterative generation. The conditional unique latent features are fixed.
    \item All conditional latent features are fixed during the entire process of iterative generation.
\end{enumerate}

\begin{figure}
    \centering
    \includegraphics[width=0.9\linewidth]{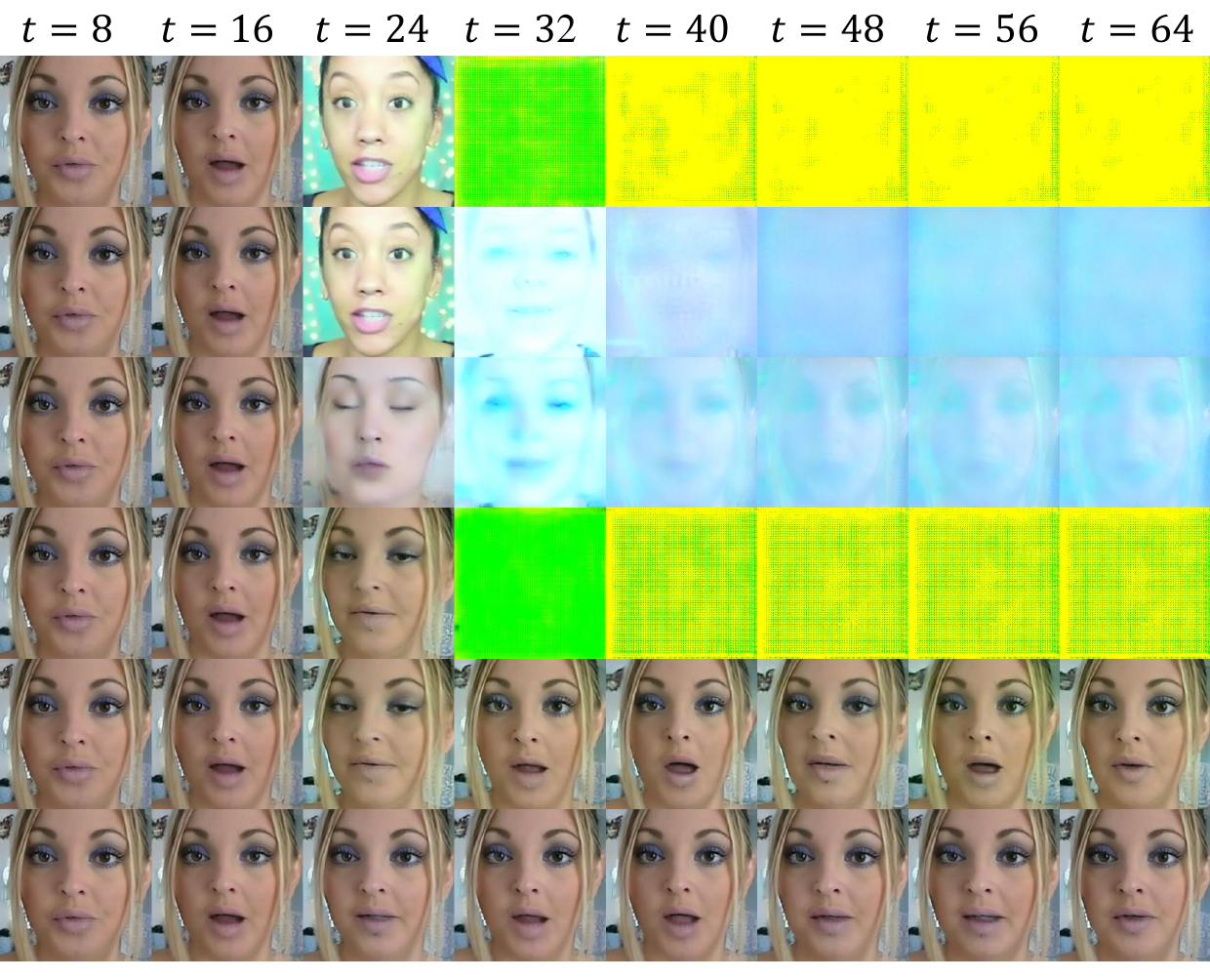}
    \caption{
    Samples generated with different sampling strategies.
    The rows from the top to the bottom depict samples generated with the 1st to the 6th sampling strategies.
    }
    \label{fig:ablate_gen}
\end{figure}
\begin{table}[t]
    \centering
    \caption{FVD scores change with different sampling strategies. Each FVD is calculated between 64 real videos and 64 generated videos. Each video contains 64 video frames.}
    \scalebox{1.2}{
        \begin{tabularx}{0.8\linewidth}{ p{0.13\linewidth}<{\centering} | 
        X<{\centering} X<{\centering} X<{\centering} X<{\centering} X<{\centering} X<{\centering} }
        \toprule
        Sampling Strategy   & 1     &   2   &   3   &   4   &   5   &   6   \\
        \midrule
        FVD                 & 7409.1& 7375.1& 7515.7& 7754.2& 6406.2& 6032.1  \\
        \bottomrule
    \end{tabularx}}
    \label{tab:ablate_ldm}
\end{table}

\begin{table*}[h]
    \centering
    \caption{Training hyper-parameters of CU-VAE.}
    \scalebox{1.2}{
        \begin{tabularx}{0.8\linewidth}{ p{0.2\linewidth}<{\centering} | 
        X<{\centering} X<{\centering} | X<{\centering} X<{\centering} }
            \toprule
            \multirow{2}{} & \multicolumn{2}{c|}{FaceForensics}    & \multicolumn{2}{c}{UCF-101}  \\
            \cline{2-5}
                                & $128^2$       & $256^2$   & $128^2$   &   $256^2$     \\
            \midrule
            Batch size              & 32        & 8         & 16        & 8     \\
            Learning Rate           & 2e-4      & 1e-4      & 1e-4      & 1e-4  \\
            Training iterations     & 200K      & 300K      & 400K      & 300K  \\
            $f_c$                   & 4         & 8         & 4         & 8     \\
            $f_u$                   & 16        & 32        & 16        & 32    \\
            $\mathcal{L}_{GAN}$     & \xmark    & \cmark    & \cmark    & \cmark    \\
            Latent embedding dim    & 3         & 3         & 3         & 3     \\
            Hidden dim              & 256       & 256       & 256       & 256   \\
            Residual layer          & 4         & 4         & 4         & 4     \\
            Discriminator hidden dim& -         & 64        & 32        & 64    \\
            \bottomrule
        \end{tabularx} }
    \label{tab:hyper-cuvae}
\end{table*}

\begin{table*}[h]
    \centering
    \caption{Training hyper-parameters of CU-LDM and ND-LDM.}
    \scalebox{1.2}{
        \begin{tabularx}{0.8\linewidth}{ p{0.2\linewidth}<{\centering} | 
        X<{\centering} X<{\centering} | X<{\centering} X<{\centering} }
            \toprule
            \multirow{2}{ } & \multicolumn{2}{c|}{FaceForensics}    & \multicolumn{2}{c}{UCF-101}  \\
            \cline{2-5}
                                    & CU-LDM    & ND-LDM    & CU-LDM    & ND-LDM    \\
            \midrule
            Batch size              & 72        & 80        & 144       & 144       \\
            Learning Rate           & 4e-5      & 4e-5      & 1e-4      & 1e-4      \\
            EMA step                & 100       & 100       & 100       & 100       \\
            Training iterations     & 250K      & 250K      & 400K      & 400K      \\
            Model dim               & 224       & 224       & 224       & 224       \\
            Attention resolutions   & [16,8,4]  & [16,8,4,2]& [16,8,4]  & [16,8,4,2] \\
            Channel multiplies      & [1,1,3,4] & [1,2,4,6] & [1,1,3,4] & [1,2,4,6] \\
            Total Parameters        & 519M      & 555M      & 519M      & 555M      \\
            \bottomrule 
        \end{tabularx} }
    \label{tab:hyper-culdm}
\end{table*}

As depicted in Fig. \ref{fig:ablate_gen}, when the given previous unique latent features are changed in the same way as the unique noise features, e.g. the 1st, 2nd and 3rd sampling strategies, the generated long videos fail to maintain realism.
Despite fixing the previous unique latent features, when the common latent feature changes in the same way as the unique noise features, e.g. the 4th sampling strategy, the generated video still contains meaningless video frames.
When we set the common latent feature to the starting one after each generation step, e.g. the 5th sampling strategy, the generated sample could maintain consistent content.
However, the sample generated by the 5th sampling strategy contains video frames with deviated coloring and inconsistent appearances, e.g. the 24th video frame is greenish.
When we fix the conditional common and unique latent features during the whole iterative generation process, i.e. the 6th sampling strategy, the generated long video keeps both content consistency and realism.
Considering that CU-LDM is trained to predict noises that have a known distribution in each step, no matter input features contain noises or not, CU-LDM always outputs noises that satisfy the known distribution, resulting in the denoising operation removing useful information of the input conditional latent features.
We also report the quantitative results of FVD on samples generated by different sampling strategies in Table \ref{tab:ablate_ldm}.
By fixing all conditional latent features, the 6th sampling strategy obtains the best FVD score.

\section*{Hyper-parameters}
We adopt linearly scheduled $\beta$ from 0.00085 to 0.0012.
When employing dpm-solver, we set the sampling order being 3 and the total sampling steps being 30.
For block-wise temporal-spatial attention calculation, matched common and unique latent features are divided into several blocks with the hyper-parameter $w$ being 2.
Additional detailed training hyper-parameters of CU-VAE, CU-LDM and ND-LDM are presented in Table \ref{tab:hyper-cuvae} and Table \ref{tab:hyper-culdm}.
Our codes will be released if accepted.

}

\vfill